\documentclass{article}

\usepackage{arxiv}

\usepackage[utf8]{inputenc} 
\usepackage[T1]{fontenc}    
\usepackage{url}            
\usepackage{booktabs}       
\usepackage{amsfonts}       
\usepackage{nicefrac}       
\usepackage{microtype}      
\usepackage{graphicx}
\usepackage{subfigure}
\usepackage{amsmath}
\usepackage{tabularx}
\usepackage{multirow}
\usepackage{hyperref}

\usepackage{colortbl}
\usepackage[inline, shortlabels]{enumitem}
\usepackage{amssymb}
\usepackage{mathtools}
\usepackage{amsthm}

\theoremstyle{definition}

\theoremstyle{remark}

\usepackage{algorithm, algorithmic, bm}
\usepackage{amsthm,amsmath,amssymb}
\usepackage{pifont}

\usepackage{wrapfig,lipsum,booktabs}

\title{Compressor-VLA: Instruction-Guided Visual Token Compression for Efficient Robotic Manipulation}

\author{
  Juntao Gao$^{1,2}$, Feiyang Ye$^{2, }$\thanks{Corresponding author}\ ~, Jing Zhang$^{1,3,*}$, Wenjing Qian$^{1}$\\
  $^1$ School of Information Science and Technology, Beijing University of Technology \\
  $^2$ LiAuto Inc. \\
  $^3$ Beijing Key Laboratory of Computational Intelligence and Intelligent System, Beijing University of Technology \\
}

\begin{document}
\maketitle

\begin{abstract}
Vision-Language-Action (VLA) models have emerged as a powerful paradigm in Embodied AI. However, the significant computational overhead of processing redundant visual tokens remains a critical bottleneck for real-time robotic deployment. While standard token pruning techniques can alleviate this, these task-agnostic methods struggle to preserve task-critical visual information. To address this challenge, simultaneously preserving both the holistic context and fine-grained details for precise action, we propose Compressor-VLA, a novel hybrid instruction-conditioned token compression framework designed for efficient, task-oriented compression of visual information in VLA models. The proposed Compressor-VLA framework consists of two token compression modules: a Semantic Task Compressor (STC) that distills holistic, task-relevant context, and a Spatial Refinement Compressor (SRC) that preserves fine-grained spatial details. This compression is dynamically modulated by the natural language instruction, allowing for the adaptive condensation of task-relevant visual information. Experimentally, extensive evaluations demonstrate that Compressor-VLA achieves a competitive success rate on the LIBERO benchmark while reducing FLOPs by 59\% and the visual token count by over 3x compared to its baseline. The real-robot deployments on a dual-arm robot platform validate the model's sim-to-real transferability and practical applicability. Moreover, qualitative analyses reveal that our instruction guidance dynamically steers the model's perceptual focus toward task-relevant objects, thereby validating the effectiveness of our approach.
\end{abstract}

\section{Introduction}
\label{introduction}

The paradigm of Vision-Language-Action (VLA) models, which directly map raw sensory inputs and language commands to low-level robot actions, represents a significant advancement in the pursuit of general-purpose robotic agents \cite{brohan2022rt, durante2025interactive}. Seminal works such as RT-1 \cite{brohan2022rt}, RT-2 \cite{zitkovich2023rt}, Gato \cite{reed2022generalist}, and more recent open-source efforts like OpenVLA \cite{kim2024openvla} and Octo \cite{team2024octo}, have demonstrated that pre-training on large-scale, diverse multimodal datasets endows these models with remarkable generalization capabilities, simplifying the traditional, decoupled ``perception-planning-control" pipeline \cite{zhou2023language, yang2025efficientvla}.

\begin{figure*}[t]
\vskip 0.2in
\begin{center}
\centerline{\includegraphics[width=\textwidth]{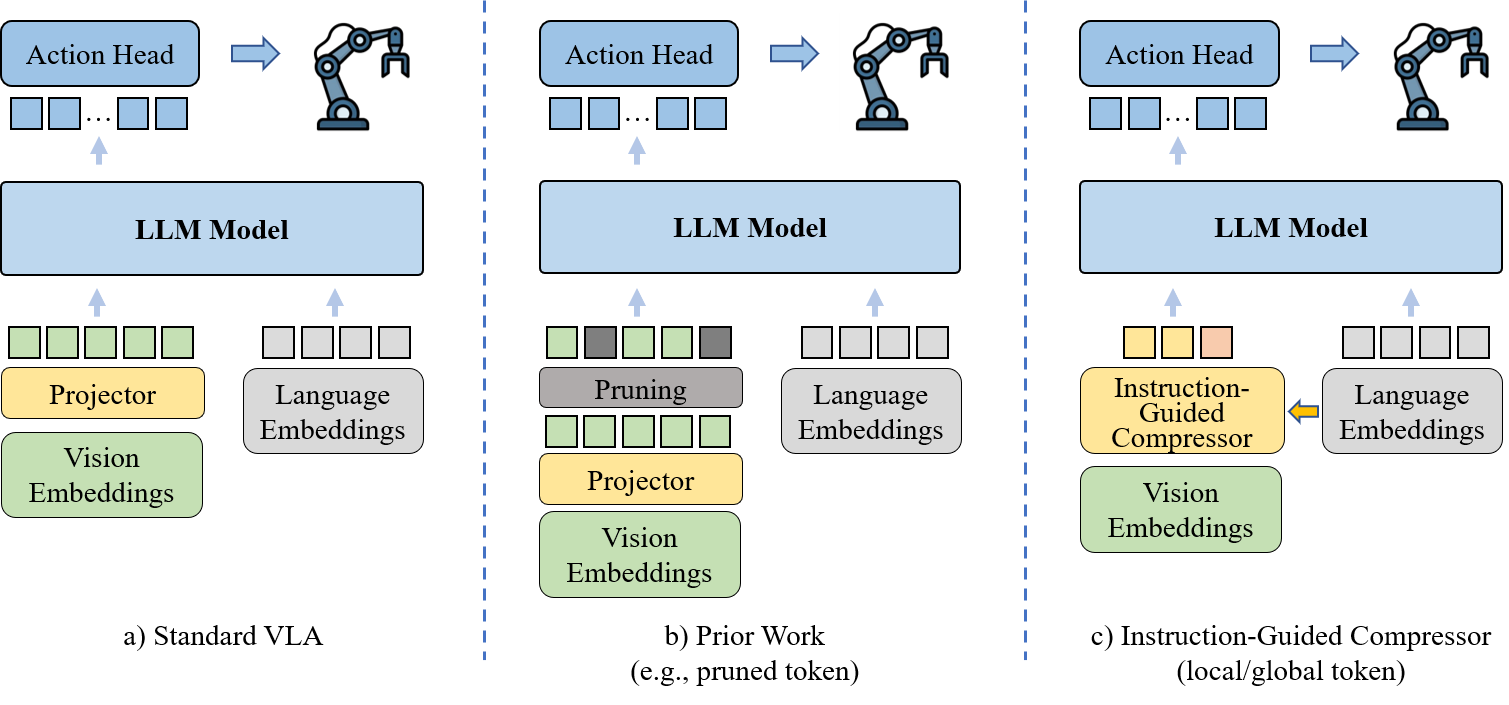}}
\caption{Comparison of three different visual information processing pipelines. (a) The standard VLA framework processes all visual tokens from the encoder. (b) Prior work, such as pruning-based methods, discards low-scoring tokens in a task-agnostic manner. (c) The proposed Compressor-VLA model performs instruction-guided compression via a dual-mechanism approach, reconstructing a compact set of tokens.}
\label{fig:pipeline}
\end{center}
\vskip -0.2in
\end{figure*}

As illustrated in Figure~\ref{fig:pipeline} (a), the architecture of these VLA models typically consists of three main core components: a high-capacity Vision Transformer (ViT) for visual perception, a Large Language Model (LLM) for high-level reasoning, and an action head to decode the LLM's output into low-level control commands. Particularly crucial is the ViT backbone, which is what grounds the model in the physical world. It deconstructs each incoming camera image into a grid of patches, which are then projected into a long sequence of hundreds or even thousands of visual tokens. This rich, high-dimensional representation encodes important information about fine-grained details and complex spatial relationships, forming the key perceptual bedrock upon which all subsequent planning and decision-making are built.

However, this reliance on a high-dimensional perceptual ability introduces a practical constraint: high computational demand. The primary reason for this overhead is the high-capacity visual stream. State-of-the-art VLAs typically employ a Vision Transformer \cite{vaswani2017attention} to encode each single image frame into a long sequence of visual tokens, as illustrated in Figure~\ref{fig:pipeline} (a). This long sequence, when processed by the subsequent LLM backbone, incurs substantial computational costs, including FLOPs and memory cost. High FLOPs will directly lead to inference latencies and make it challenging for real-time interaction \cite{huang2025otter} in robot manipulation. Furthermore, these raw visual inputs also contain lots of task-irrelevant information (e.g., background clutter, complex textures), which both consumes computational resources and can introduce noise that degrades policy quality \cite{zhang2025beyond, yang2025topv}. 

To address this efficiency challenge, a common approach is token reduction, which typically involves a pruning layer that discards tokens based on task-agnostic importance scores \cite{zhang2025beyond, yang2025topv, alvar2025divprune}. While effective at reducing token count, this ``hard" selection strategy risks losing crucial information. More importantly, this entire approach fails to leverage the language instruction to guide the compression process. This is suboptimal for robotics, where the relevance of visual cues is highly dynamic and dependent on the given instruction. Other related methods include token merging \cite{bolya2022tome} and query-based compression \cite{alayrac2022flamingo}, but they often share the same task-agnostic limitation.

In this work, we propose Compressor-VLA, a novel token compression framework for VLA models, as illustrated in Figure~\ref{fig:pipeline} (c). Instead of directly pruning tokens in previous works, our key insight is using a hybrid instruction-guided compressor to \textit{reconstruct} a compact set of tokens. The proposed Compressor-VLA framework achieves this through a dual-mechanism approach where both pathways are modulated by the language instruction: the \textbf{Semantic Task Compressor (STC)} employs \textit{cross-attention compression} to distill a holistic, task-oriented scene summary, and the \textbf{Spatial Refinement Compressor (SRC)} uses \textit{guided local attention} to preserve essential spatial details and topology. The experiments show that this reconstructive and instruction-guided design enables more efficient and context-aware compression. Our main contributions are as follows:
\begin{itemize}[leftmargin=5.5mm]
    \item We propose the Compressor-VLA, a novel hybrid instruction-conditioned token compression framework that combines both local and global information pathways, designed for efficient, task-oriented compression of visual information in VLA models. 
    \item Extensive experiments on the challenging LIBERO benchmark demonstrate that the proposed Compressor-VLA framework achieves a superior efficiency-performance trade-off, reducing FLOPs by 59\% while achieving a competitive 97.3\% success rate. The results of real-robot deployments on a dual-arm robot platform also validate the effectiveness of our approach. 
    \item We provide qualitative and quantitative analyses that validate the core design principles of our framework, revealing the effectiveness of the instruction-guidance token compression strategies and the architectural synergy between the global and local compression pathways.
\end{itemize}

\section{Related Work}

\paragraph{Vision-Language-Action Models.}
Drawing from the success of pretrained vision, LLM, and VLM foundations, VLAs have emerged to process multimodal inputs-visual observations and language instructions to generate robotic actions for embodied tasks. RT-1 \cite{brohan2022rt} and Octo \cite{reed2022generalist} utilize a transformer-based policy integrating diverse data, such as robot trajectories across varied tasks, objects, environments, and embodiments. Recent studies \cite{zitkovich2023rt, kim2024openvla, livision} harness pretrained VLMs to produce robotic actions, tapping into world knowledge from large-scale vision-language datasets. For example, RT-2 \cite{zitkovich2023rt} and OpenVLA \cite{kim2024openvla} model actions as tokens within the language vocabulary, whereas RoboFlamingo \cite{livision} adds a separate policy head for action prediction. OpenVLA \cite{kim2024openvla} and Octo \cite{team2024octo}, trained on large, aggregated datasets such as Open X-Embodiment \cite{vuong2023open}, have become crucial benchmarks for the community. To mitigate catastrophic forgetting and better leverage the powerful priors of VLMs, methods like OTTER \cite{huang2025otter} freeze the language-aligned vision encoders, such as CLIP \cite{radford2021learning} and train a lightweight policy network on top of text-aware visual features. While architectural choices differ, all these models face the common challenge of efficiently processing long visual token sequences.

\paragraph{Efficient Token Processing}
One way to optimize VLM inference is to reduce the visual tokens that occupy the majority of the input sequence. Several studies have explored efficient token processing in language models \cite{dai2020funnel, huang2022pyramid}. Compared with text, image information tends to have higher redundancy, making visual token reduction for VLMs more reasonable and effective. Several strategies have been made to compress visual tokens into a more compact representation, such as token pruning \cite{zhang2025beyond, yang2025topv}, merging \cite{bolya2022tome}, and query-based aggregation \cite{alayrac2022flamingo}. Most previous efficient Token Processing methods in VLA employ a pruning method to reduce the length of the visual token sequence, such as VLA-Cache \cite{xu2025vla} and EfficientVLA \cite{yang2025efficientvla}. However, our work is closely related to query-based aggregation, which uses a fixed set of queries as an information bottleneck. Moreover, unlike most prior work that is task-agnostic, our entire compression process is task-conditioned, ensuring that it is a goal-oriented filtering process rather than a generic summarization. 

\paragraph{Instruction-Conditioned Mechanisms}
Modulating a network's behavior based on an external context is a powerful technique, ranging from high-level cross-attention \cite{vaswani2017attention} to fine-grained feature modulation. The latter, exemplified by methods like Feature-wise Linear Modulation (FiLM) \cite{perez2018film} and Adaptive Instance Normalization (AdaIN) \cite{huang2017arbitrary}, uses a conditioning signal to generate affine transformation parameters (scale and shift) that are applied to a network's intermediate representations. In our work, we employ FiLM to design a lightweight and efficient instruction-conditioned mechanism, using the language instruction to guide the visual compression.

\section{Method}


\subsection{Overall Architecture and Data Flow}
\begin{figure*}[t]
\vskip 0.2in
\begin{center}
\centerline{\includegraphics[width=\textwidth]{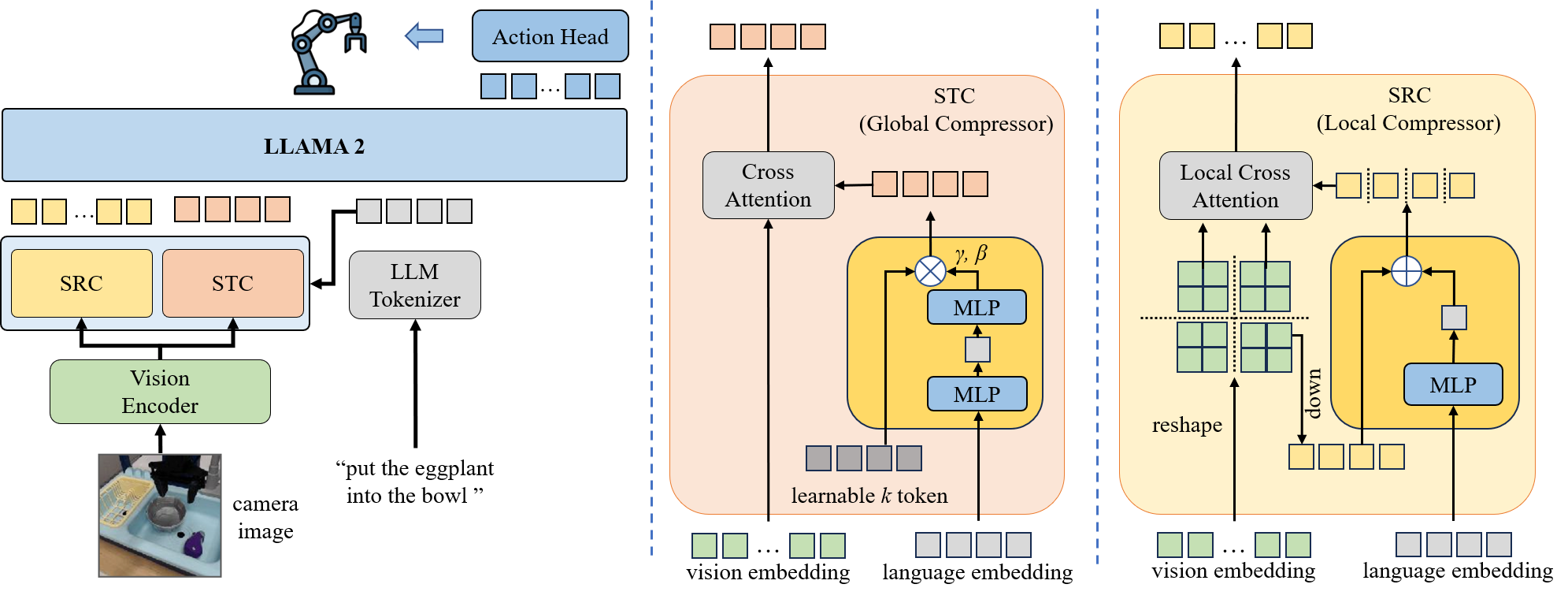}}
\caption{The architecture of the proposed Compressor-VLA. The module features two instruction-guided parallel pathways. The Semantic Task Compressor (STC) uses language to modulate its queries, while the Spatial Refinement Compressor (SRC) infuses language information directly into local visual tokens.}
\label{fig:arch}
\end{center}
\vskip -0.2in
\end{figure*}

To address the efficiency bottleneck in VLA models, we introduce a hybrid instruction-guided compressor. This token compression module replaces the standard projector connecting the vision encoder and the LLM backbone, as illustrated in Figure~\ref{fig:arch} (c).

The proposed token compression module operates on two primary inputs: the visual features $X \in \mathbb{R}^{N \times D}$ from a pre-trained vision encoder, where $N$ and $D$ represent the number of vision tokens and embedding dimension, respectively, and the language instruction embeddings from the VLA's own LLM backbone. To accommodate variable-length instructions, we apply mean pooling across the sequence of language embeddings to derive a single fixed-size vector, $L_{pooled}$. This vector subsequently functions as the conditioning signal for instruction guidance. We do not use a separate, pre-trained language encoder (e.g., CLIP) for two reasons. First, this choice reduces the model's overall parameter count, contributing to a more lightweight design. Second, the LLM's internal text embeddings, trained on vast corpora, are posited to contain sufficiently rich semantic information to serve as an effective modulation signal.

The data flows through two parallel pathways. This hybrid architecture can balance two competing and equally critical requirements in robotic manipulation: high-level semantic understanding and low-level spatial precision. A separate global compressor, similar to standard Perceiver-style models, excels at creating a holistic scene summary but risks removing the fine-grained spatial details necessary for precise actions. The proposed token compression model resolves this challenge by creating a hybrid architecture, where the two components handle different complementary tasks:

\begin{itemize}[leftmargin=5.5mm]
    \item The \textbf{Semantic Task Compressor (STC)} provides the strategic context (\textit{what to do and where to go}) by generating a globally-contextualized summary $Z_G \in \mathbb{R}^{k \times D}$.
    \item The \textbf{Spatial Refinement Compressor (SRC)} supplies the tactical precision (\textit{how to act}) by producing a spatially-aware local representation $Z_L \in \mathbb{R}^{N' \times D}$.
\end{itemize}
The final compressed visual token sequence is composed of the outputs of STC and STC, i.e., $Z = \text{Concat}([Z_G; Z_L])$. This compressed visual token sequence is then input into the LLM backbone to generate the robot's action. The specific design of STC and SRC is introduced in detail in the following sections.

\subsection{Semantic Task Compressor}
The goal of the STC is to distill the visual scene into a compact, fixed-length summary that is dynamically conditioned on the language instruction. It employs a query-based aggregation strategy, using a small set of $k$ learnable queries, $Q \in \mathbb{R}^{k \times D}$, where $k \ll N$. We modulate these queries using Feature-wise Linear Modulation (FiLM) \cite{perez2018film}. Specifically, a semantic representation of the task, denoted by $E_L$, is computed by passing the pooled language vector through a dedicated MLP: 
\begin{align}
    E_L = \text{MLP}_{\text{STC}}(L_{pooled})
\end{align}
Then, a small MLP generates per-query affine transformation parameters (scale $\gamma$ and shift $\beta$):
\begin{align}
    &\gamma, \beta = \text{MLP}_{\text{FiLM}}(E_L) \\
&Q_{\text{con}} = \gamma \odot Q + \beta
\end{align}
These conditioned queries then interact with the full set of visual tokens $X$ via cross-attention to produce the global summary $Z_G$:
\begin{equation}
Z_G = \text{Attention}(Q= Q_{\text{con}}, K=X, V=X)
\end{equation}
Our approach ensures that the task instruction from the very beginning shapes the information bottleneck. Since the STC's learnable queries can act as abstract ``concept detectors." We employ FiLM, as it provides a powerful, non-linear way to adapt these detectors based on the instruction, effectively asking, ``for this task, which concepts should I be looking for in the scene?"

\subsection{Spatial Refinement Compressor}
To compensate for the potential spatial information loss in the global pathway, the goal of SRC is to reduce token count while preserving fine-grained, task-relevant local details. It operates on non-overlapping local windows of the 3D feature $X' \in \mathbb{R}^{H \times W \times D}$, where $H$, $W$, and $D$ represent the height, width, and embedding dimension, respectively.

For each window containing tokens $X_w \in \mathbb{R}^{w\times w \times D}$, we first generate a representative ``raw" query, denoted by $q_{raw}$, by downsampling and reshaping the original visual features. This query is then modulated by the language instruction. A transformed instruction embedding is produced from the pooled language vector via a separate MLP, $E'_L = \text{MLP}_{\text{SRC}}(L_{pooled})$, and added to the raw query. This conditioned query, denoted by $q_w$, then attends to the original, unmodified window tokens to produce a task-aware summary $z_w$:
\begin{align}
    q_{raw} &= \text{Downsample}(X_w) \\
q_w &= q_{raw} + E'_{L} \\
z_w &= \text{Attention}(Q=q_w, K=X_w, V=X_w)
\end{align}
The operator $\text{Downsample}(\cdot)$ conducts downsampling and flattens to the input vector. The final local representation $Z_L$ is the concatenation of all $z_w$. This design efficiently infuses task-relevance into the query process, guiding the model to preserve local details pertinent to the instruction. For the SRC, the primary goal is to preserve local detail with minimal distortion. We therefore opt for a simpler direct injection of the language embedding into the query. This linear shift acts as a gentle ``hint" to the local query, subtly biasing its attention towards task-relevant features within its window without aggressively transforming the representation, thereby safeguarding the high-fidelity spatial information required for precise manipulation.

\section{Experiments} 
To demonstrate the performance of our proposed Compressor-VLA framework, we conduct experiments on a diverse suite of settings, including simulation tasks and real-world scenarios. Additionally, we conduct a comprehensive ablation study and qualitative analysis to validate the effectiveness of our approach.

\subsection{Simulation Tasks}
\paragraph{Simulation Environment} We conduct simulation experiments on the LIBERO benchmark \cite{liu2023libero}, which features a Franka Emika Panda arm in simulation with demonstrations containing camera images, robot state, task annotations, and delta end-effector pose actions. Specifically, we use the tasks and datasets in LIBERO-Spatial, LIBERO-Object, LIBERO-Goal, and LIBERO-Long, which contain diverse objects, scene layouts, and language instructions. Each task suite consists of 10 tasks with 50 human-teleoperated demonstrations.

\paragraph{Baselines}
We evaluate Compressor-VLA against two types of VLA methods:
(1) General-purpose VLAs: We use OpenVLA-OFT \cite{kim2025fine} as our primary baseline. Since it acts as the base architecture, this comparison isolates the specific impact of our compression mechanisms. Additionally, we compare against representative, high-performing models, including CogACT \cite{li2024cogact} and $\pi0$ \cite{black2024pi_0}.
(2) Efficient VLAs: We benchmark against five pruning-based approaches, including SP-VLA \cite{li2025sp}, FastV \cite{chen2024image}, SparseVLM \cite{zhang2024sparsevlm}, VLA-Cache \cite{xu2025vla}, and SpecPrune-VLA \cite{wang2025specprune}.

\paragraph{Evaluation Metrics} For the experiments on the LIBERO benchmark, we employ standard performance evaluation metrics, i.e., the percentage of trials where the task is completed successfully, calculated by $\text{Success Rate} = \frac{\text{Successful episodes}}{\text{Total episodes}} \times 100\%$. In addition, we evaluate the model efficiency by FLOPs, the floating-point operations calculated by the ``thop" library, and the compressed token count, the number of tokens after compression. For those adaptive pruning-based baselines, such as SP-VLA and SpecPrune-VLA, we report their average compressed token.

\paragraph{Implementation Details}
All experiments are conducted on 8 Nvidia A100 GPUs. For the proposed Compressor-VLA framework, we use the open-sourced OpenVLA-OFT \cite{kim2025fine} as our foundation model, which consists of a two-branch vision encoder (DINOv2 and SigLIP) and a LLaMA-2-7B backbone. Following \cite{kim2025fine}, the model is initialized from the official OpenVLA checkpoints. To fine-tune the model, we apply the LoRA technique \cite{hu2022lora} with a rank of 32 to the vision encoder and LLM backbone, while the action head and proprioceptive projector and two proposed compressors are fully fine-tuned. The model is trained for 150,000 gradient steps with an initial learning rate of 5e-4, which includes a warm-up phase from 10\% of the value for stability. The learning rate is decayed to 5e-5 after 100,000 steps. We use a batch size of 8 per device for a global batch size of 64. We set the default number of the global queries $k=16$ and the size of local window $w=2$. Following \cite{wang2025specprune}, we report the results of VLA-Cache methods on the OpenVLA-OFT model to do a fair comparison, since the original method was developed on the OpenVLA model.

\begin{table*}[!th]
\caption{Performance on LIBERO-Spatial, LIBERO-Object, LIBERO-Goal, and LIBERO-Long task suites. The best results for each task on each measure over all methods are highlighted in \textbf{bold}. MIN. indicates the minimum required token numbers.}
\label{tab:main_results}
\vskip 0.15in
\begin{center}
\begin{small}
\begin{tabular}{lccccccc}
\toprule
\multirow{2}{*}{Model} & Spatial  & Object  & Goal  & Long  & Avg.  & FLOPs & \multirow{2}{*}{Compressed Tokens} \\
& SR (\%) & SR (\%) & SR (\%) & SR (\%) & SR (\%) & (T) &  \\
\midrule
\multicolumn{8}{l}{\textit{General-Purpose VLA Models}} \\
OpenVLA-OFT \cite{kim2025fine} & 97.6 & 98.4 & 97.9 & 94.5 & 97.1 & 3.95 & 512 \\
CogACT \cite{li2024cogact} & 97.2 & 98.0 & 90.2 & 88.8 & 93.6 & - & 512 \\
$\pi0$ \cite{black2024pi_0} & 96.8 & 98.8 & 95.8 & 85.2 & 94.2 & - & 512 \\
\midrule
\multicolumn{8}{l}{\textit{Efficient VLA Methods}} \\
SP-VLA \cite{li2025sp} & 75.4 & 85.6 & 84.4 & 54.2 & 74.9 & 3.10 & 229 (avg.) \\
FastV \cite{chen2024image} & 96.8 & 81.0 & 96.4 & 73.0 & 86.8 & 3.18 & 256 \\
SparseVLM \cite{zhang2024sparsevlm} & 96.8 & 94.2 & 97.6 & 93.6 & 95.6 & 3.04 & 256 \\
SpecPrune-VLA \cite{wang2025specprune} & 98.2  & 96.3 & \textbf{97.7} & 94.0 & 96.6 & 1.70 & 197 (avg.) \\
VLA-Cache \cite{xu2025vla} & \textbf{99.0} & 97.7 & 97.4 & 93.6 & 97.0 & 3.28 & 212 (min.) \\
\midrule
\textbf{Compressor-VLA} & 98.8 & \textbf{99.2} & 96.4 & \textbf{94.8} & \textbf{97.3} & \textbf{1.62} & 160 \\
\bottomrule
\end{tabular}
\end{small}
\end{center}
\vskip -0.1in
\end{table*}

\paragraph{Simulation Results}
The main experimental results on the LIBERO benchmark are reported in Table~\ref{tab:main_results}.

The primary finding is that Compressor-VLA achieves a good balance between efficiency and performance. Compared to the OpenVLA-OFT baseline, our model reduces FLOPs by 59\% (from 3.95T to 1.62T) and compresses the token sequence by over 3x (from 512 to 160), while maintaining a highly competitive average success rate (97.1\% vs. 97.3\%). This demonstrates the effectiveness of Compressor-VLA in preserving critical task-relevant information during compression. Furthermore, when compared to other general-purpose VLA models like CogACT and $\pi0$, our model achieves a significantly higher success rate with a much lower computational cost.

Compared with other efficient VLA methods, our method still shows advantages in both FLOPs and effectiveness. Take VLA-Cache for example, it caches the similar and unimportant tokens' key and value from the last generation and reuses them in the next inference. Therefore, it does not significantly reduce computational costs. Therefore, although they can also achieve relatively high experimental results, their FLOPs are not superior to ours. The SpecPrune-VLA method has lower FLOPs, but its success rate is not competitive enough. Therefore, these results show that the proposed framework offers a better balance between high performance and computational efficiency.

\subsection{Ablation Study}
\label{sec:ablation}
To dissect the contribution of each component in our Compressor-VLA, we conduct ablation studies on both the compressor components, instruction guidance, and hyperparameters. 

\begin{table*}[!t]
\caption{Ablation on compressor components and guidance on LIBERO-Spatial, LIBERO-Object, LIBERO-Goal, and LIBERO-Long task suites. The best results for each task on each measure are highlighted in \textbf{bold}.}
\label{tab:ablation_components}
\vskip 0.15in
\begin{center}
\begin{small}
\begin{tabular}{lccccccc}
\toprule
\multirow{2}{*}{Model Configuration} & Spatial  & Object  & Goal  & Long  & Avg.  & FLOPs & \multirow{2}{*}{Compressed Tokens} \\
& SR (\%) & SR (\%) & SR (\%) & SR (\%) & SR (\%) & (T) &  \\
\midrule
STC+SRC & \textbf{98.8} & \textbf{99.2} & 96.4 & \textbf{94.8} & \textbf{97.3} & 1.62 & 160 \\
STC+SRC-FiLM & 98.4 & 99.0 & 96.0 & 94.4 & 97.0 & 1.62 & 160 \\
No Guidance & 98.6 & 99.0 & 93.8 & 93.8 & 96.3 & 1.43 & 160 \\
STC-Only & 96.0 & 97.6 & \textbf{96.8} & 93.0 & 95.9 & \textbf{0.76} & \textbf{32} \\
SRC-Only & 97.6 & 98.2 & 94.2 & 92.2 & 95.5 & 1.20 & 128 \\
\bottomrule
\end{tabular}
\end{small}
\end{center}
\vskip -0.1in
\end{table*}

\subsubsection{Compressor Components and Instruction Guidance}
To study the contributions of each architecture component in Compressor-VLA, we conduct additional experiments on the LIBERO benchmark with five different model configurations: \textbf{STC+SRC} represents the original model setting of Compressor-VLA. \textbf{STC+SRC-FiLM} represents adding FiLM to the SRC model. \textbf{No Guidance} means removing the instruction information injection in both STC and SRC models. \textbf{STC-Only} means removing the instruction information injection in the SRC model. \textbf{SRC-Only} means removing the instruction information injection in the STC model.

The results of five model settings are reported in Table~\ref{tab:ablation_components}. The results validate our design choices. The proposed `STC+SRC' architecture achieves the highest success rate of 94.8\%. The importance of this dual-pathway structure is confirmed by the significantly lower performance of the `STC-Only' (93.0\%) and `SRC-Only' (92.2\%), and `No Guidance' models, demonstrating that both global context and local detail are important. In addition, the `STC+SRC-FiLM' variant yielded a slightly lower success rate (94.4\%) compared with `STC+SRC', suggesting that the simpler direct injection is more effective for preserving the SRC's fine-grained spatial information.

\subsubsection{Sensitivity to Hyperparameters}
We conduct the model's sensitivity analysis on the two key hyperparameters, the number of global queries ($k$) in the STC, and the local window size ($w$) in the SRC. The setting of these two hyperparameters controls the visual token compression trade-off between the compactness of the global summary and the granularity of spatial detail preservation.

We study the choice of these two hyperparameters on the LIBERO-Long task suite, and the results are shown in Table~\ref{tab:ablation_hyperparams}. For global queries ($k$), performance peaks at $k=16$ and $k=32$. We choose $k=16$ as our default, as it provides the same top performance with lower computational cost (1.62T FLOPs vs. 1.83T). For the local window size $w$, a size of $w=2$ yields the best result (94.8\%). As the window size increases, performance drops, confirming that smaller windows are crucial for preserving the fine-grained spatial details necessary for manipulation tasks. These results show that the choice of the number of global queries is less sensitive than the choice of the local window size, and validate our default hyperparameter choices.

\begin{table}[th]
\caption{Ablation on hyperparameters $k$ and $w$ on the LIBERO-Long task suite. The best results on each measure of each group are highlighted in \textbf{bold}.}
\label{tab:ablation_hyperparams}
\vskip -0.15in
\begin{center}
\begin{small}
\begin{tabular}{lcccc}
\toprule
\multirow{2}{*}{Hyperparam.} & \multirow{2}{*}{Value} & Long & FLOPs  & \multirow{2}{*}{Tokens} \\
&  & SR (\%) & (T) &\\
\midrule
\multirow{3}{*}{Global Queries} & $k=8$ & 94.2 & \textbf{1.51} & \textbf{144} \\
 & $k=16$ & \textbf{94.8} & 1.62 & 160 \\
 & $k=32$ & \textbf{94.8} & 1.83 & 192 \\
\midrule
\multirow{3}{*}{Local Window} & $w=2$ & \textbf{94.8} & 1.62 & 160 \\
 & $w=4$ & 92.6 & 0.86 & 64 \\
 & $w=8$ & 91.4 & \textbf{0.70} & \textbf{40} \\
\bottomrule
\end{tabular}
\end{small}
\end{center}
\vskip -0.1in
\end{table}

\subsection{Real-World Tasks}
To validate the sim-to-real transferability and practical applicability of our method, we deployed Compressor-VLA on a real-world Mobile ALOHA robot. To evaluate our framework, we design two tasks to test the model's ability on spatial awareness and semantic understanding, respectively.

\paragraph{Datasets}
We perform evaluations on two challenging task: (1) \textbf{Spatial Awareness}: Place the bucket in the center and put the objects to be picked up (simulated toys, yellow banana, green pepper, purple eggplant) into the bucket (``put X into bucket"). (2) \textbf{Semantic Understanding}: Stack the medium tower on top of the large one first, then stack the small one on top of the medium one.(``Stack tower of hanoi").

For the first task, we collect 100 trajectories for each object. During testing, the buckets are placed in fixed positions, while the objects are randomly placed. For the second task, we collected 10 trajectories for each combination of relative positions, and test the model in a similar relative placement.

\paragraph{Implementation Details}
All real-world experiments are conducted with a cobot magic dual-arm robot. This robot is equipped with two identical Piper arms from AgileX Robotics, resulting in a total of 14 degrees of freedom (7-DoF for each arm). For visual input, a third-view Orbecc DABAI RGB-D camera is used, from which we only utilize the RGB images. The model is initialized from the official OpenVLA checkpoints and then finetuned for 100,000 gradient steps. Consistent with our simulation experiments, we use a learning rate of 5e-4, which is decayed to 5e-5 after 50,000 steps.

\begin{figure*}[t]
\begin{center}
\centerline{\includegraphics[width=0.9\textwidth]{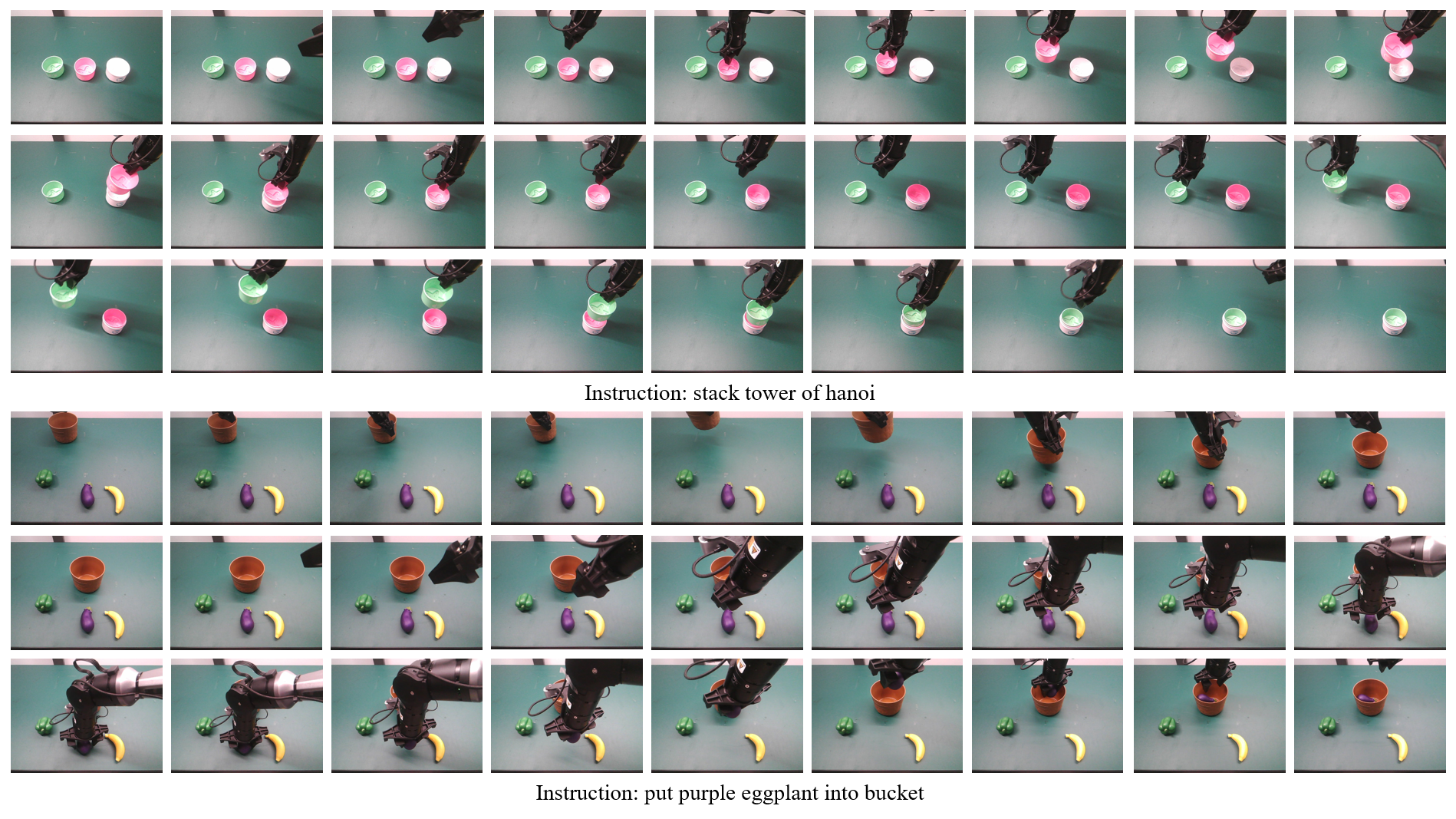}}
\caption{Execution examples on real-world tasks.}
\label{fig:hanoi2}
\end{center}
\vskip -0.2in
\end{figure*}

\paragraph{Results}
Execution examples on real-world tasks are shown in Figure \ref{fig:hanoi2}, and the experimental results are reported in Table~\ref{tab:real_world}. These results show that Compressor-VLA has both spatial awareness and semantic understanding capabilities. In addition, the proposed Compressor-VLA achieves a high success rate, comparable to the original OpenVLA-OFT baseline, demonstrating successful transfer from simulation to the real world.

\begin{table}[th]
\caption{Real-World task success rate on Mobile ALOHA.}
\label{tab:real_world}
\vskip -0.15in
\begin{center}
\begin{small}
\begin{tabular}{lcc}
\toprule
\multirow{2}{*}{Model} & Spatial Awareness & Semantic Understanding \\
 & (SR \%) & (SR \%) \\
\midrule
OpenVLA-OFT & 91.7 (22/24) & 76.7 (23/30) \\
Compressor-VLA & \textbf{100} (24/24) & 83.3 (25/30) \\
\bottomrule
\end{tabular}
\end{small}
\end{center}
\vskip -0.1in
\end{table}

\clearpage

\subsection{Qualitative Analysis}
\label{sec:qualitative}
To provide intuitive insights into the working mechanism of the proposed Compressor-VLA framework, we present qualitative analyses that validate our two core design principles: instruction-guided compression and the effect of the STC and SRC architecture.

\subsubsection{Instruction-Conditioned Attention}
\begin{figure*}[!t]
\vskip -0.2in
\begin{center}
\centerline{\includegraphics[width=0.9\textwidth]{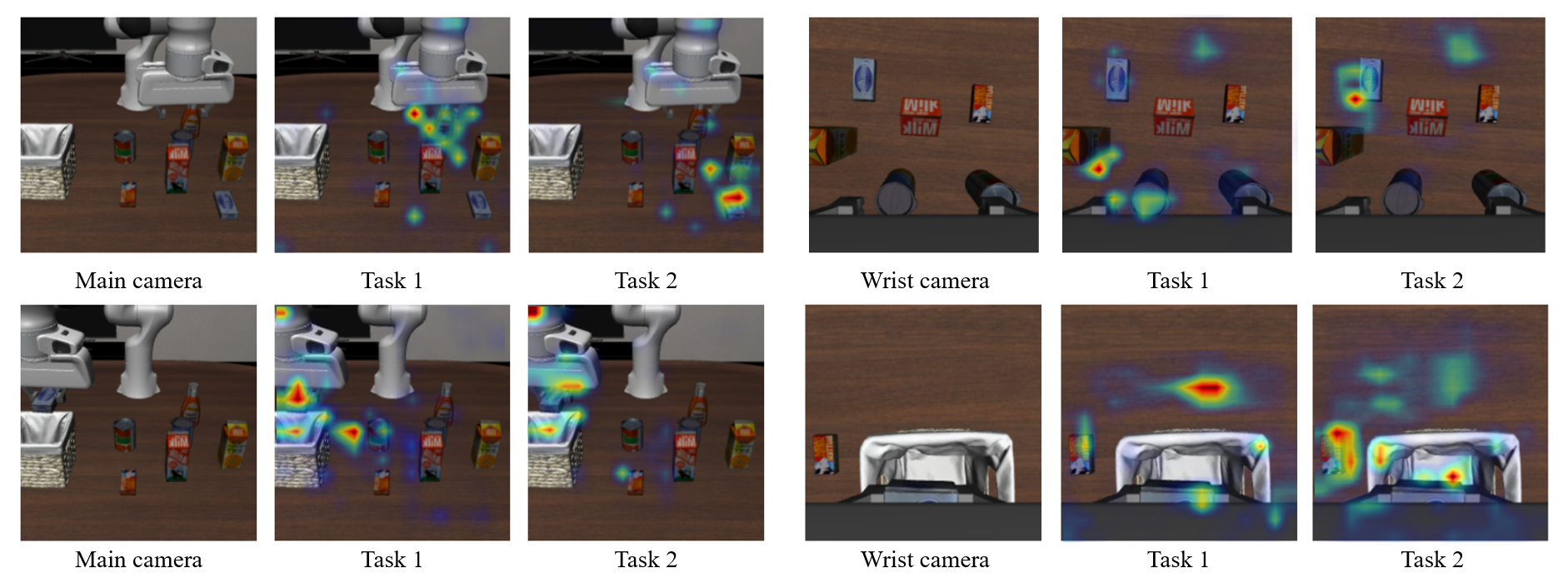}}
\caption{Instruction-conditioned attention visualization. The STC's module attention with the same initial scene and different language commands. Left: ``put both the alphabet soup and the tomato sauce...". Right: ``put both the cream cheese box and the butter...".}
\label{fig:attention}
\end{center}
\vskip -0.2in
\end{figure*}

One core goal of our work is to ensure that the compression process is dynamically guided by the language instruction. To verify this, we visualize the attention maps from the STC's queries for the identical initial scene from `LIVING ROOM' under two different multi-object commands. The visualization result is shown in Figure~\ref{fig:attention}, and it shows our task-oriented focusing mechanism.

When the command is ``put both the alphabet soup and the tomato sauce in the basket," the model's attention probes correctly focus on the soup can, which is the first object to be grasped. Critically, when the command changes to ``put both the cream cheese box and the butter in the basket," the attention probes dynamically shift to the cream cheese box-again, the next immediate target. This demonstrates two key findings: (1) The instruction guidance mechanism effectively directs the model's perceptual resources to the correct target objects based on the task description. (2) The compressor exhibits a temporally-aware focusing behavior, inherited from the sequential nature of VLA pre-training. It does not waste capacity by attending to all the mentioned objects simultaneously, but intelligently anticipates and concentrates on the most imminent sub-goal. This validates that our module performs not just compression, but efficient, goal-directed, and predictive information filtering. Furthermore, a consistent, low-level attention is maintained on the robot's own manipulator across tasks, indicating a stable proprioceptive awareness that is crucial for control.

\begin{figure*}[!t]
\vskip -0.2in
\begin{center}
\centerline{\includegraphics[width=0.9\textwidth]{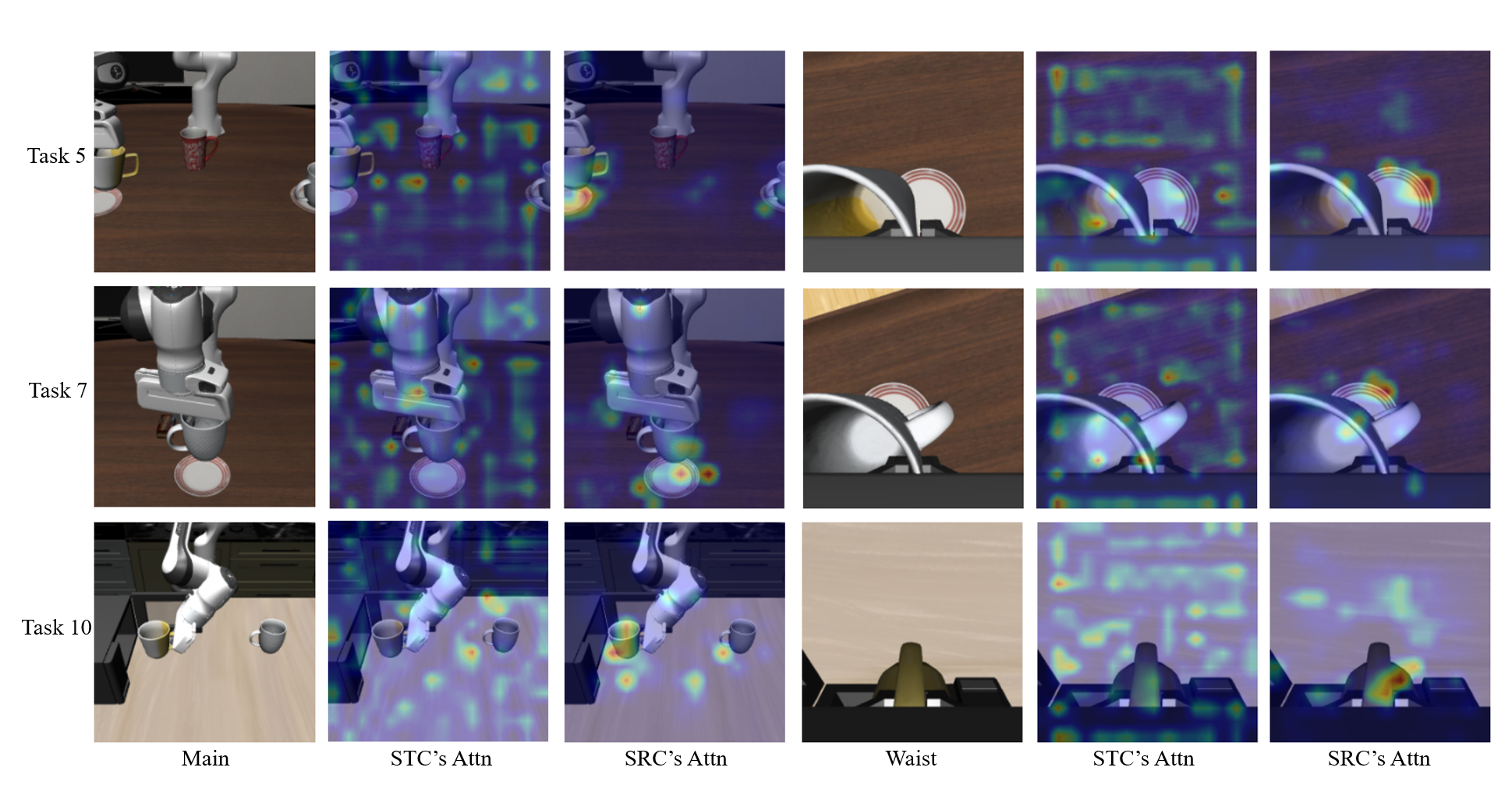}}
\caption{Visualization of the hybrid architecture's synergy across multiple tasks in LIBERO-10, including Task 5 (``put the white mug on the left plate..."), Task 7 (``put the white mug on the plate..."), and Task 10 (``put the yellow and white mug in the microwave..."). }
\label{fig:synergy}
\end{center}
\vskip -0.2in
\end{figure*}

\subsubsection{Hybrid Architecture}
To validate the complementary roles of the STC and SRC, we visualize their attention outputs across a diverse set of challenging tasks from the LIBERO-10 benchmark, as shown in Figure~\ref{fig:synergy}. The results demonstrate a synergistic division of labor among these two modules.

The STC model consistently acts as a high-level planner. It processes the wide-angle view to identify and locate key objects relevant to the instruction. For instance, in Task 5 (``put the white mug on the left plate..."), its attention correctly focuses on the two distinct mugs, which is the crucial first step for disambiguating the complex spatial command. This demonstrates its strength in understanding the overall scene and forming a high-level plan of \textit{what} to interact with and \textit{where} the targets are.

In contrast, the SRC specializes in preserving the fine-grained details necessary for precise physical interaction. While its attention appears more distributed due to its windowed operation, it exhibits the ability to highlight critical sub-object features. For example, in Task 10 (``put the yellow and white mug in the microwave..."), its attention focuses on the mug's handle, a crucial detail for a stable grasp. Similarly, in Task 7 (``put the white mug on the plate..."), it focuses on the rim of the mug, which is essential for generating the correct action chunk.

\section{Conclusions}
We propose the Compressor-VLA framework to address the critical efficiency bottleneck in Vision-Language-Action models. Our framework performs instruction-guided visual token compression by combining a Spatial Refinement Compressor for spatial detail and a Semantic Task Compressor for contextual summary. Extensive experiments on simulation tasks demonstrate that Compressor-VLA significantly reduces computational load while maintaining competitive task success rates. Experiments on real-world tasks further validate the sim-to-real transferability and practical applicability of our method. We show the effectiveness of our hybrid architecture and validate our design choices through ablation studies. Our work validates that by intelligently filtering visual data based on task relevance at an early stage, we can achieve more efficient and robust robotic agents.

\bibliographystyle{unsrt}

\end{document}